\documentclass[12pt, a4paper]{article}
\usepackage[margin=1in]{geometry}
\usepackage{amsmath, amssymb, graphicx, hyperref, listings}
\usepackage{graphicx}
\usepackage{epstopdf}
\usepackage{algorithm,algorithmic}
\usepackage{caption}
\usepackage{subcaption}
\usepackage{indentfirst}

\graphicspath{{figures/}}

\begin{document}

\title{Sketch 'n Solve: An Efficient Python Package for Large-Scale Least Squares Using Randomized Numerical Linear Algebra}

\author{
    Alex Lavaee$^{1}$ \
    \texttt{alavaee@bu.edu} \
}

\date{}

\maketitle
\begin{center}
    \textsuperscript{1}Boston University \
    \vspace{0.2cm}
    \textit{May 2023}
\end{center}

\begin{abstract}
    We present Sketch 'n Solve, an open-source Python package that implements efficient randomized numerical linear algebra (RandNLA) techniques for solving large-scale least squares problems. While sketch-and-solve algorithms have demonstrated theoretical promise, their practical adoption has been limited by the lack of robust, user-friendly implementations. Our package addresses this gap by providing an optimized implementation built on NumPy and SciPy, featuring both dense and sparse sketching operators with a clean API. Through extensive benchmarking, we demonstrate that our implementation achieves up to 50x speedup over traditional LSQR while maintaining high accuracy, even for ill-conditioned matrices. The package shows particular promise for applications in machine learning optimization, signal processing, and scientific computing.
\end{abstract}

\section{Introduction}

\subsection{Context: Randomized Numerical Linear Algebra}
Randomized numerical linear algebra (RandNLA) has emerged as a powerful approach for developing efficient algorithms for large-scale linear algebra computations. As described by Murray et al. \cite{murray2023randomized}, RandNLA originated from the insight that randomization can compute approximate solutions to linear algebra problems more efficiently than deterministic algorithms. This approach has proven particularly valuable in machine learning and statistical data analysis applications.

The integration of RandNLA with numerical analysis and classical numerical linear algebra has led to its incorporation into widely-used software packages including MATLAB, the NAG Library, NVIDIA's cuSOLVER, and scikit-learn. One of the fundamental techniques in RandNLA is matrix sketching, which reduces problem dimensionality while preserving essential structure.

\subsection{The Sketch-and-Solve Paradigm}
The sketch-and-solve paradigm consists of three steps:

\begin{enumerate}
    \item Generate a sketch matrix $\mathbf{S}$
    \item Create a sketched matrix $\mathbf{SA}$ and sketched vector $\mathbf{Sb}$
    \item Solve the computational problem on $\mathbf{SA}$ using existing algorithms
\end{enumerate}

The choice of sketching matrix $\mathbf{S}$ is crucial, as it determines both the approximation quality and computational complexity. Sketching operators can be categorized into two main types: dense and sparse.

\subsubsection{Dense Sketching Operators}
Dense sketching operators generate matrices where most entries are non-zero. While theoretically easier to analyze, they can be computationally expensive for large-scale problems. The primary dense operators in common use are Gaussian sketches \cite{woodruff2014sketching}, which sample entries independently from a Gaussian distribution, and Hadamard sketches \cite{tropp2011improved}, which are constructed using normalized Hadamard matrices and efficiently computed using Fast Hadamard Transform.

\subsubsection{Sparse Sketching Operators}
Sparse sketching operators generate matrices where most entries are zero, offering improved computational efficiency through reduced memory access and arithmetic operations. The most notable sparse operators are the Clarkson-Woodruff sketches \cite{woodruff2014sketching}, which combine a sparse matrix with a random sign matrix offering strong theoretical guarantees, and sparse sign embeddings \cite{Martinsson_Tropp_2020}, which are constructed by randomly selecting and sign-modifying matrix rows or columns.

Our experimentation showed that sparse sketching operators consistently outperformed dense ones, with the Clarkson-Woodruff sketch ultimately chosen as the default sketching operator in our implementation due to its theoretical guarantees and superior performance.

\begin{figure}[H]
    \centering
    \includegraphics[scale=0.8]{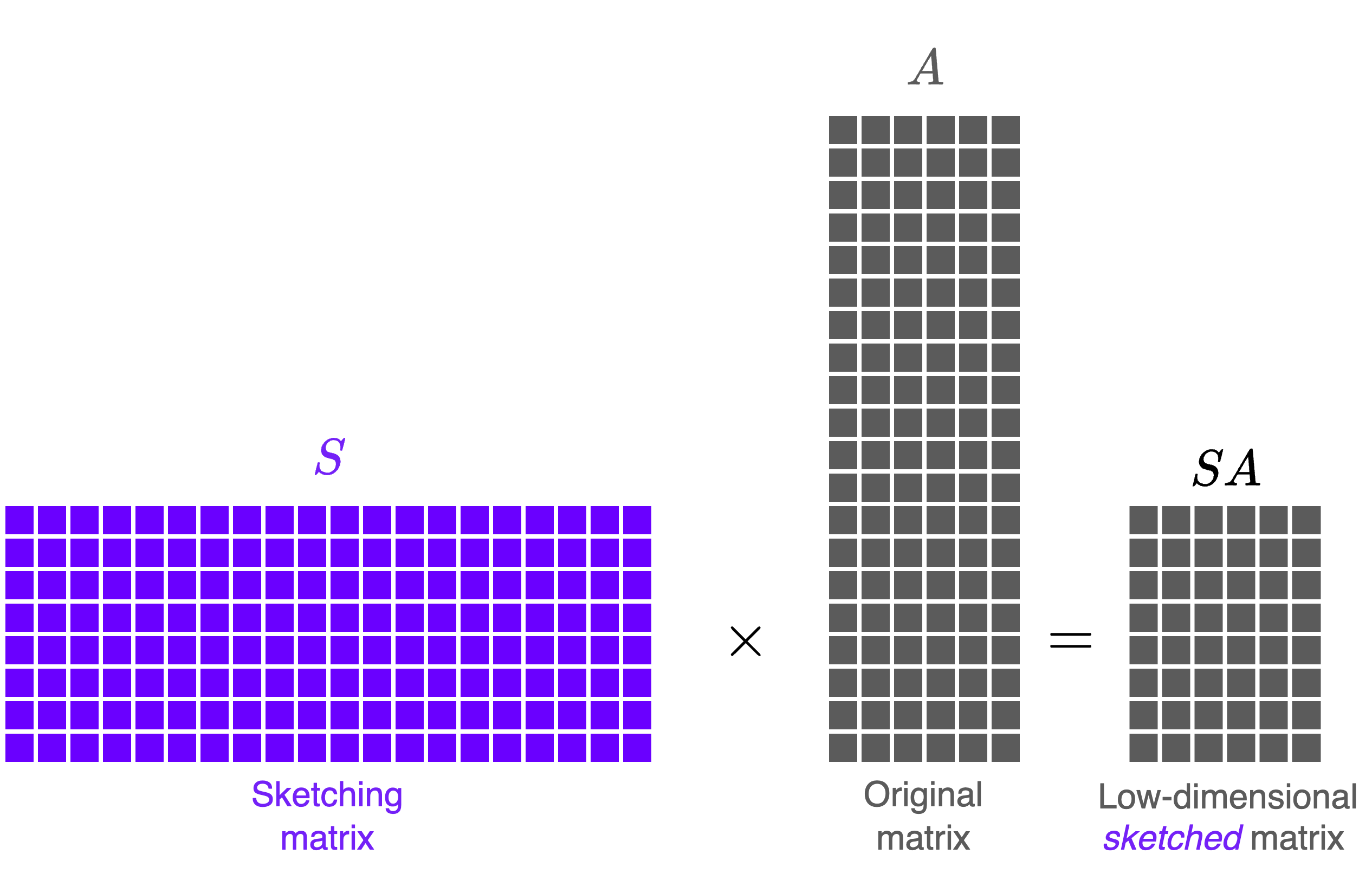}
    \caption{Dense sketch matrix}
\end{figure}

\begin{figure}[H]
    \centering
    \includegraphics[scale=0.8]{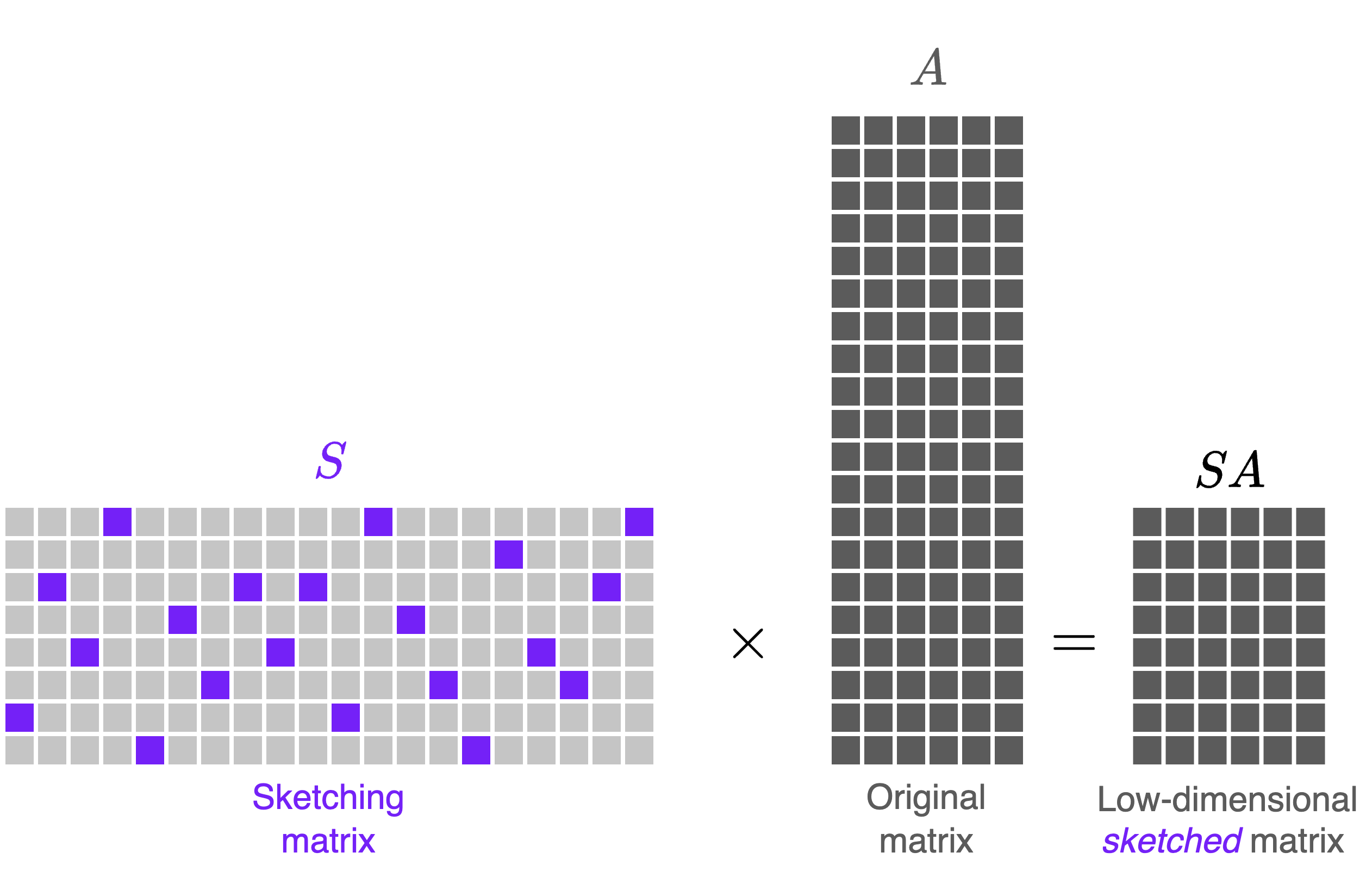}
    \caption{Sparse sketch matrix}
\end{figure}

\subsection{The Least Squares Problem}
The overdetermined least squares objective is:

\begin{align}
    \min_{\mathbf{x}} \| \mathbf{A}\mathbf{x} - \mathbf{b}\|^2 \label{eq:least_squares} \
    \text{for } \mathbf{A} \in \mathbb{R}^{m \times n}, \mathbf{x} \in \mathbb{R}^n, \mathbf{b} \in \mathbb{R}^m
\end{align}

In sketch-and-solve, we find a sketch matrix $\mathbf{S} \in \mathbb{R}^{d \times m}$ ($m \gg n \approx d$) to solve the reduced problem:

\begin{align*}
    \min_{\hat{\mathbf{x}}} \|\mathbf{SA}\hat{\mathbf{x}} - \mathbf{Sb}\|^2
\end{align*}

\subsection{Motivation and Contributions}
Despite their theoretical advantages, sketch-and-solve algorithms have faced several adoption challenges. These include the absence of robust, production-ready implementations, complex interfaces requiring deep RandNLA understanding, limited integration with scientific computing ecosystems, and insufficient support for both dense and sparse matrices.

Our work makes several key contributions to address these challenges. First, we provide a robust Python implementation optimized for both dense and sparse sketch matrices. Second, we develop an intuitive API that makes RandNLA techniques accessible to a broader audience. Finally, we ensure seamless integration with the NumPy/SciPy ecosystem, allowing researchers and practitioners to easily incorporate our methods into existing workflows.

\section{Implementation}

\subsection{API Design}
Sketch 'n Solve builds on NumPy and SciPy's optimized implementations of fundamental linear algebra operations. The package architecture comprises three main components. First, the sketching operators component implements both dense and sparse sketching. Second, the linear system solvers component provides efficient sketch-and-solve algorithms using both the sketch and precondition (SAP-SAS) and sketch and apply (SAA-SAS) algorithms \cite{meier2023sketchandprecondition}. Third, the utility functions component handles sketch matrix creation and manipulation.

The implementation features several key characteristics that enhance its usability and performance. The package provides a unified interface for different sketching operators, automatically handles dense and sparse matrices, and integrates seamlessly with NumPy and SciPy array and sparse matrix interfaces.

\subsection{Optimization Techniques}
Our implementation incorporates several optimization strategies to enhance performance. The package leverages efficient sparse matrix operations through SciPy, employs optimized sketching operators using BLAS routines, carefully manages memory for large-scale problems using sparse matrices, and utilizes vectorized operations for computational efficiency.

\section{Experimental Results}

\subsection{Benchmark Setup}
Following \cite{epperly2024fast}, we work with ill-conditioned matrices sampled from a standard normal distribution, with dimensions $m$ ranging from $5 \times 10^3$ to $10^6$ and $n = 10^3$. We set the condition number $\kappa = 10^{10}$ and residual norm parameter $\beta = 10^{-10}$.

\subsection{Performance Analysis}
We compare our implementation against SciPy's LSQR implementation. Figures 3-5 show runtime and error comparisons.

\begin{figure}[H]
    \centering
    \includegraphics[scale=0.75]{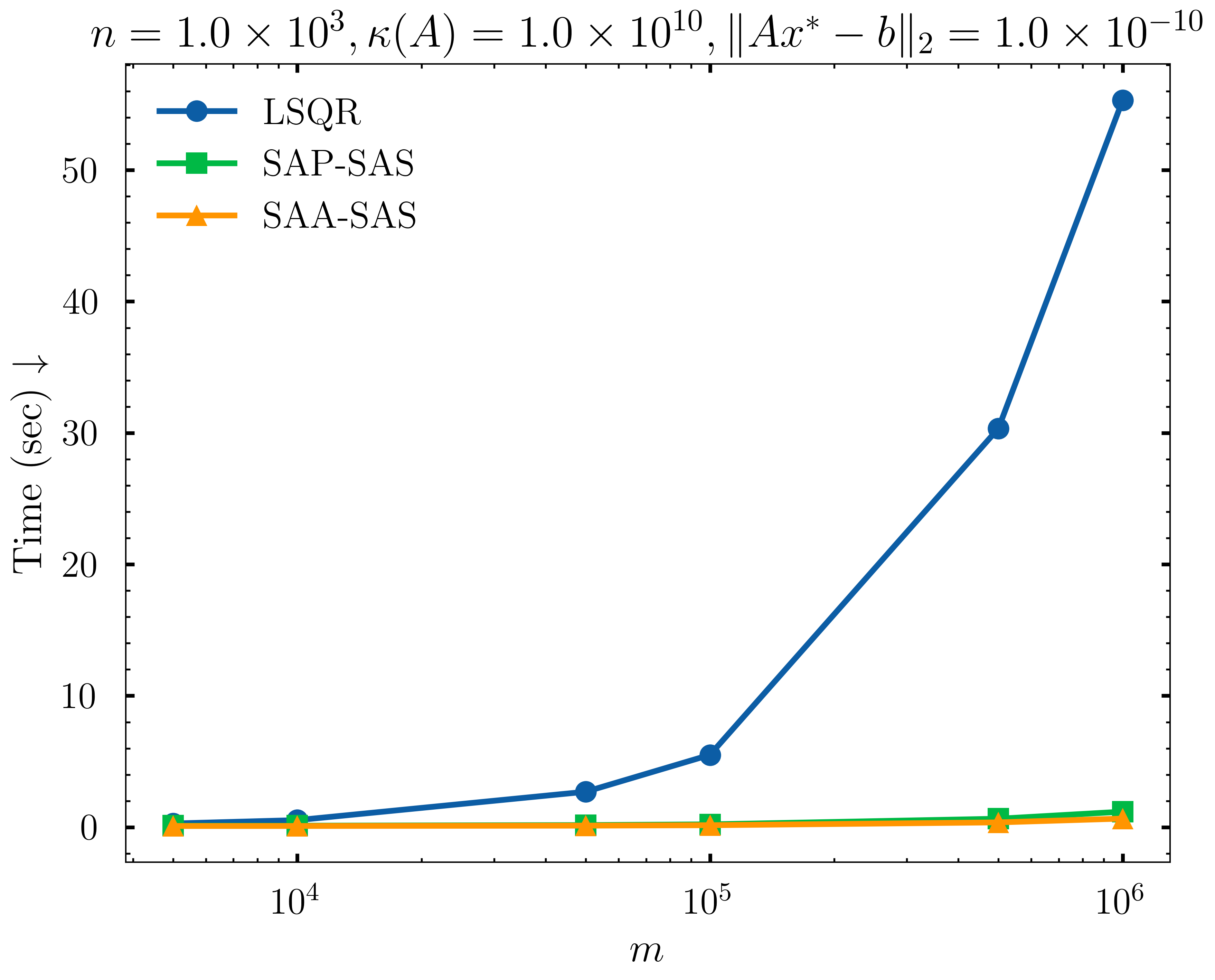}
    \caption{Runtime comparison between SAP-SAS, SAA-SAS, and LSQR (baseline).}
\end{figure}

\begin{figure}[H]
    \centering
    \includegraphics[scale=0.75]{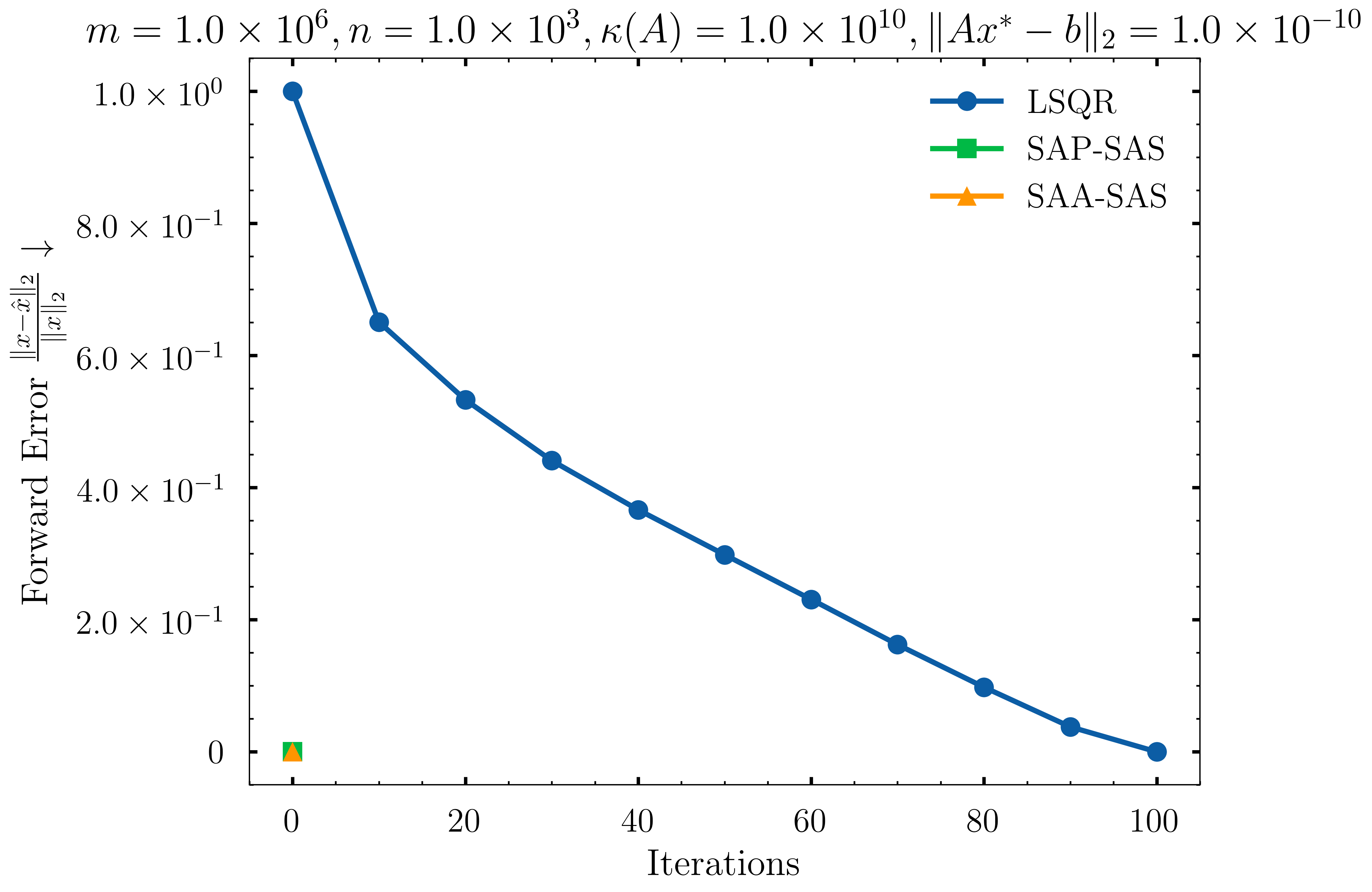}
    \caption{Forward error comparison between SAP-SAS, SAA-SAS, and LSQR (baseline).}
\end{figure}

\begin{figure}[H]
    \centering
    \includegraphics[scale=0.75]{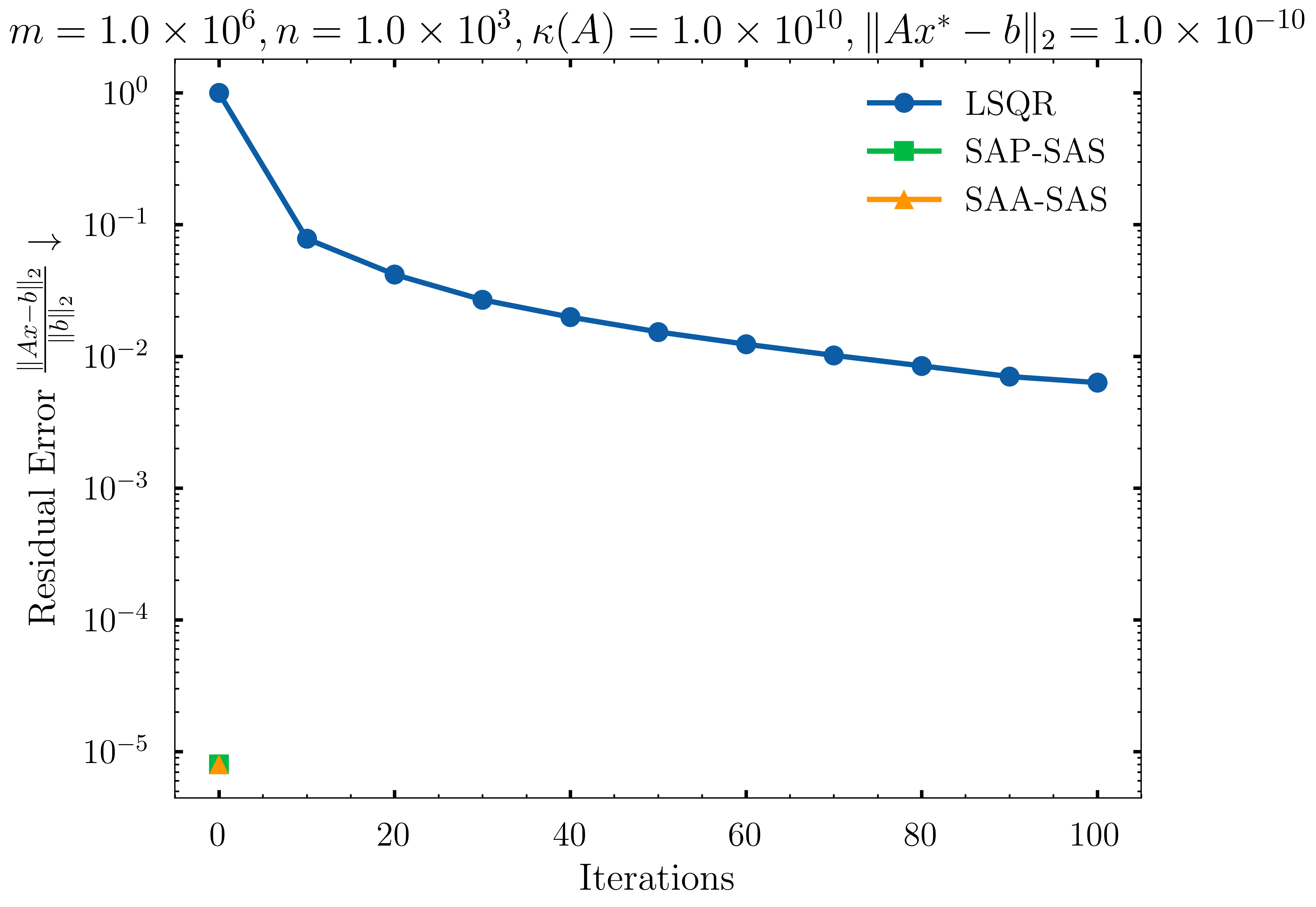}
    \caption{Residual error comparison between SAP-SAS, SAA-SAS, and LSQR (baseline).}
\end{figure}

Our results demonstrate that our implementation consistently outperforms LSQR in runtime while maintaining improved accuracy especially for ill-conditioned matrices, with the performance gap between the runtime widening as matrix size increases.

\subsection{Applications}
Sketch 'n Solve has demonstrated utility across various domains. In machine learning, the package has proven valuable for large-scale regression problems, neural network training optimization, and high-dimensional data preprocessing. In scientific computing, it has found applications in numerical simulation and modeling, signal processing and image reconstruction, and statistical data analysis.

\section{Code Availability}
The package is available at \url{https://github.com/lavaman131/sketch-n-solve}, containing implementation code and experimental reproduction materials.

\section{Conclusion}
Sketch 'n Solve makes efficient RandNLA techniques accessible through a clean, intuitive API. Our implementation demonstrates significant performance improvements over traditional methods while maintaining high accuracy, even for ill-conditioned problems.

Several directions for future work have been identified. These include integration with additional scientific computing frameworks such as PyTorch and TensorFlow, support for distributed computing implementations, and the development of additional sketching operators to further enhance the package's capabilities.

\bibliographystyle{plain}
\bibliography{refs}

\end{document}